\documentclass[letterpaper, 10 pt, conference]{ieeeconf}  

\IEEEoverridecommandlockouts                              

\usepackage{graphicx} 
\usepackage{color}
\usepackage{multicol}
\usepackage{amsmath} 
\usepackage{amssymb}  
\usepackage{balance}
\usepackage[colorinlistoftodos,prependcaption,textsize=tiny,color=red!15]{todonotes}

\usepackage{subcaption}

\usepackage{comment}
\excludecomment{hide}

\presetkeys{todonotes}{inline}{}

\title{\LARGE \bf
``The World Is Its Own Best Model'':\\Robust Real-World Manipulation Through Online Behavior Selection
}

\author{Manuel Baum$^{1,2}$ \qquad Oliver Brock$^{1,2}$ 
  \thanks{$^1$ Robotics and Biology Laboratory, Technische Universit\"at Berlin}
  \thanks{$^2$ Science of Intelligence (SCIoI), Cluster of Excellence, Berlin, Germany}
  \thanks{We gratefully acknowledge funding by the Deutsche Forschungsgemeinschaft (DFG, German Research Foundation) under Germany's Excellence Strategy -- EXC 2002/1 ``Science of Intelligence'' -- project number 390523135.}
}

\begin{document}

\bibliographystyle{IEEEtran}

\maketitle
\thispagestyle{empty}
\pagestyle{empty}

\begin{abstract}

Robotic manipulation behavior should be robust to disturbances that violate high-level task-structure. Such robustness can be achieved by constantly monitoring the environment to observe the discrete high-level state of the task. This is possible because different phases of a task are characterized by different sensor patterns and by monitoring these patterns a robot can decide which controllers to execute in the moment. This relaxes assumptions about the temporal sequence of those controllers and makes behavior robust to unforeseen disturbances. We implement this idea as probabilistic filter over discrete states where each state is direcly associated with a controller. Based on this framework we present a robotic system that is able to open a drawer and grasp tennis balls from it in a surprisingly robust way.

\end{abstract}


\section{INTRODUCTION}

In his paper \emph{Elephants don't play chess}~\cite{brooks_elephants_1990} Rodney Brooks stated that ``the world is its own best model. It is always exactly up to date. It always contains every detail there is to be known. The trick is to sense it appropriately and often enough''. It is a powerful idea to replace error-prone modeling of state variables by online perception, whenever it is possible. This has been successfully applied to extract properties of objects and geometry, for example, in the exploitation of environmental constraints~\cite{eppner_ijrr_2015} using compliant manipulators~\cite{deimel16-IJRR}. But how far can we go with this idea? Can we also offload more complex properties to the environment?

Complicated tasks usually require robots to solve several sub-tasks in sequence. Roboticists have thus developed a variety of different structures to arrange controllers into a solution for the overarching task. Each of these structures implements other assumptions about tasks and behavior.

When task solutions are planned and explicitly represented as a sequence, this implements temporal assumptions about the evolution of the task. Although linear plans are easy to design, they are likely to fail when unforeseen events occur. To resolve this, contingency plans~\cite{pryor_planning_1996} make plan execution conditional on such events, but they quickly become complex and may still not capture all contingencies.  Often we do not need to anticipate all events though and we can instead select controllers online, based on the state of the environment.

Another assumption that can be made about behavior is that it is structured in a hierarchical way. A famous implementation of this assumption is the subsumption architecture~\cite{brooks_robust_1986}. In this architecture controllers are ordered hierarchically so that higher level behaviors subsume lower levels. However, the assumption that one behavior is more high-level than another is sometimes completely artificial~\cite{hartley91}.

Both, temporal and hierarchical assumptions are made to resolve ambiguities, in case multiple controllers are applicable in the same state. But do we really need to make such assumptions? In this paper we show that the multimodal sensor input in two typical robotic manipulation settings is informative enough to directly choose controllers online -- without temporal or hierarchical assumptions.  The behavior in both tasks relies on interactive perception~\cite{bohg_interactive_2017} to actively boost the information content of sensor input. This way we can show that even a simple manipulation system using standard tools from control and state estimation can yield behavior, which is extremely robust to severe disturbances.

\begin{figure}
 \includegraphics[width=\linewidth]{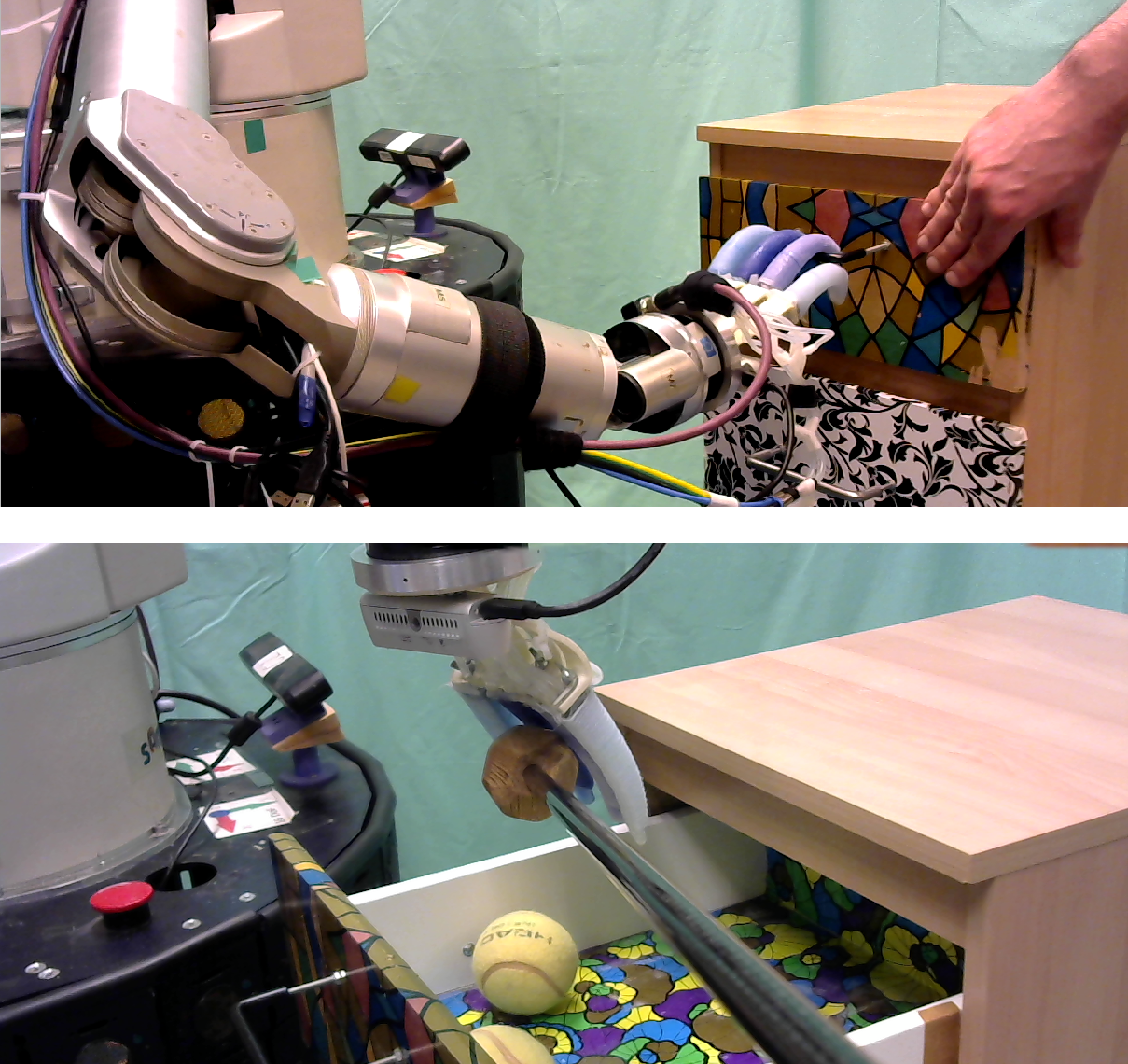}
 \caption{A robot opens a drawer and grasps tennis balls from it even while a human deliberately disturbs the robot during the execution of these tasks. This is enabled by a reactive method that does not assume that a certain sequence of controllers solve that task. Instead this sequence is constructed online, based on feedback.}
 \label{fig:robot}
\end{figure}

\section{RELATED WORK}

Robotic behavior that switches between controllers to solve sub-tasks needs to ensure that the active controller fits the current discrete state of the environment. We will now discuss approaches to select controllers for execution. We first discuss approaches that rely on predictions to make such decisions a-priori and then discuss approaches that increasingly use feedback. We evaluate them by how robust the behavior is to forseeable and, more importantly, \emph{unforeseeable} disturbances.

Standard planning approaches assume a determinstic world and that it is possible to predict what the state of the world is after an action~\cite{russell2002artificial}. But in practice the real world not deterministic, so linear plans are likely to fail under disturbances and behavior will not be robust. 

Other approaches soften that assumption so that the state may transition to one out of several states. For example, contingency plans are more flexible, tree-like plans~\cite{pryor_planning_1996} that may also be implemented as hybrid automata~\cite{egerstedt_behavior_2000}. But as these only consider a limited set of temporal evolutions and transitions, they are only robust to \emph{foreseeable} disturbances. They fail if contingencies were not anticipated. Attempts to capture all possible contingencies quickly lead to excessively complex graphs. We avoid complexity by perceiving transitions instead of predicting them.

Some approaches make even fewer assumptions about the temporal evolution of tasks. Aspect Transition Graphs (ATG)~\cite{ku_object_2015} filter the discrete state and allow arbitrarily switches in the state estimate, however task-directed behavior is achieved by continuously replanning on that discrete state and actions are partially executed open-loop~\cite{ku_aspect_2017}. This presumably limits their robustness, although online state estimation lends the approach some robustness against disturbances.  Our work also constantly monitors the world but continuously performs feedback control to be more robust.

Reactive planning methods aim to fully avoid predictions about how the discrete state of a task evolves and instead act conditional on the environment~\cite{firby_investigation_1987}. The MOSAIC model~\cite{haruno_mosaic_2001} can be seen as a reactive planning method. It switches between different competing behaviors based on forward models and responsibility predictors attached to controllers. That controller whose predictors best explain the current sensor context is then executed. MOSAIC can arbitrarily switch between controllers and is potentially robust against disturbances. It is closely related to our approach, however we abstain from using predictive forward models and perform Bayesian filtering on the state.

Two approaches that are very robust to disturbances are the subsumption architecture~\cite{brooks_intelligence_1991} and Robust Logical-Dynamical Systems (RLDS)~\cite{paxton_representing_2019}.
Subsumption architecture~\cite{brooks_intelligence_1991} decomposes behavior hierarchically into layers where high-level behaviors subsume lower levels. Subsumption architecture does not assume temporal structure, but assumes a hierarchical ordering of behaviors to resolve ambiguities. In practice, behavior often cannot be decomposed into sub-behaviors with \emph{hierarchical} ordering\cite{gat_three-layer_1998} and \emph{horizontal} competition between behaviors is required as in our approach. Similarly, Robust Logical-Dynamical Systems (RLDS)~\cite{paxton_representing_2019} achieve remarkable robustness to high-level disturbances, but also assume an ordering similar to subsumption architecture. Additionally RLDS rely on a logic state estimate, which is not always possible and necessary. We largely avoid that problem and directly use sensor data to decide which controllers to activate. We use state-estimation only where it is absolutely necessary.






\section{Task-Phase Estimation Instead of Temporal Assumptions}\label{sec:method:stateestimation}

We will now explain how we use state estimation instead of temporal assumptions in linear plans. After this general explanation we describe the concrete instantiations for drawer opening and grasping tennis balls in Section~\ref{sec:method:instantiation}.

\begin{figure}
\vspace{0.2cm}
 \includegraphics[width=\linewidth]{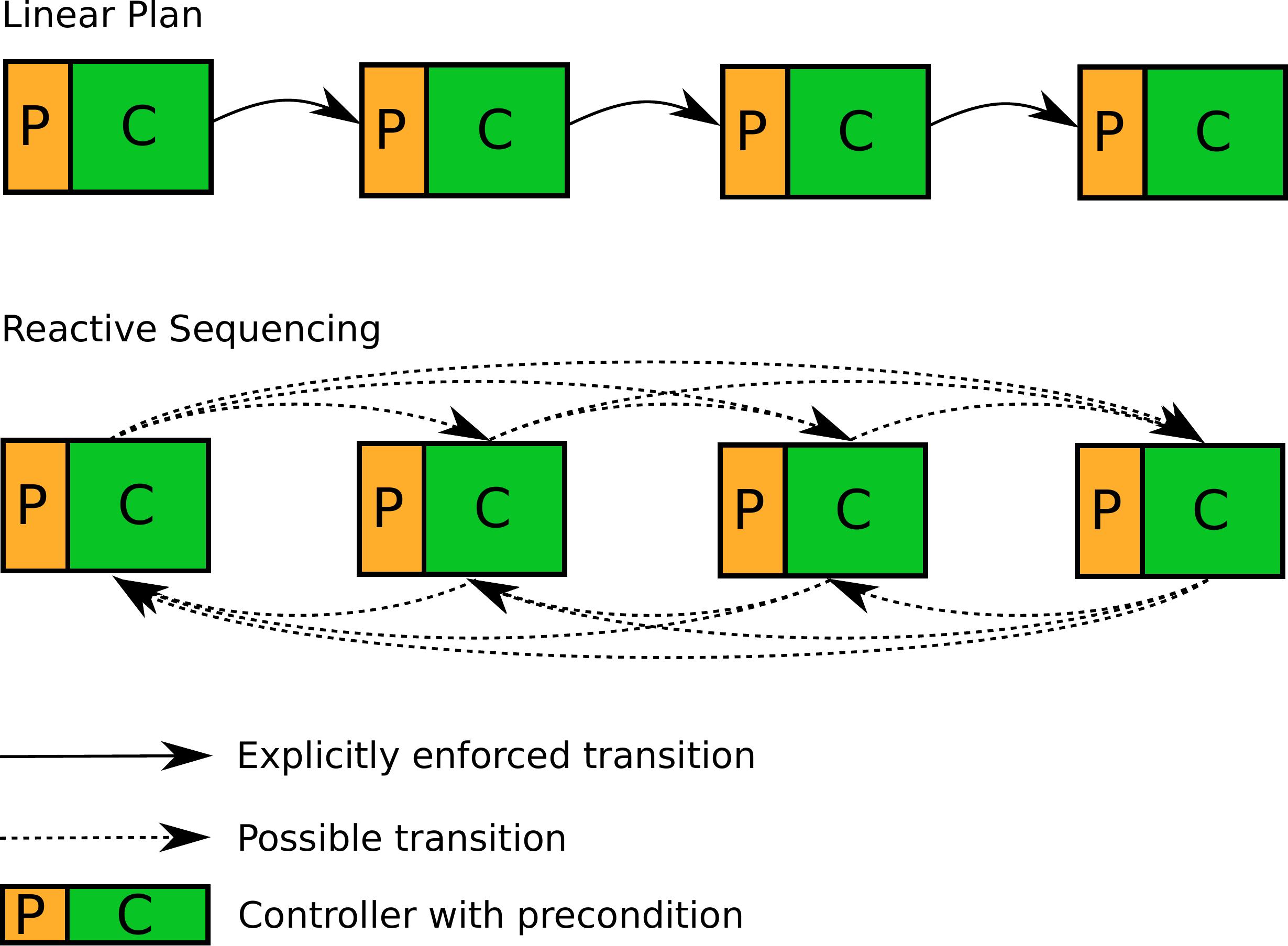}
 \caption{This shows a comparison between linear plans (top) and our approach (bottom). We remove the explicitly enforced transitions between nodes and instead allow arbitrary transitions between nodes.}\label{fig:method:comparison}
\vspace{-0.25cm}
\end{figure}

Complex manipulation tasks often require robots to solve sub-problems in different phases of the task. Each phase often requires a different controller to be executed and this poses the problem to decide which controller to activate at each point in time. One approach to solve this problem is to a-priori create a linear plan as in the top part of Figure~\ref{fig:method:comparison}, where pairs of controllers and their pre-conditions are linked in sequence. To decide if the behavior should switch to the next controller, at runtime only the precondition of the directly subsequent controller needs to be checked. This scheme can so simple because it assumes a rigid temporal structure. Although these assumptions make sequential plans prone to failure if unanticipated disturbances occur, squential plans have the advantage that they are simple to understand and design. For these reasons we first sketched such a linear plan for each of our experiments, but then transfered those sketches to the following representation, that is more robust to unanticipated disturbances.

Robust behavior should be able to react to arbitrary transitions between states, as visualized in the lower part of Figure~\ref{fig:method:comparison}. This requires to anticipate transitions to each discrete state at the same time. A standard tool to solve this problem is filtering by using Hidden Markov Models (HMM) and variants~\cite{kroemer_learning_2014}. 

\begin{figure}
\vspace{0.2cm}
 \includegraphics[width=\linewidth]{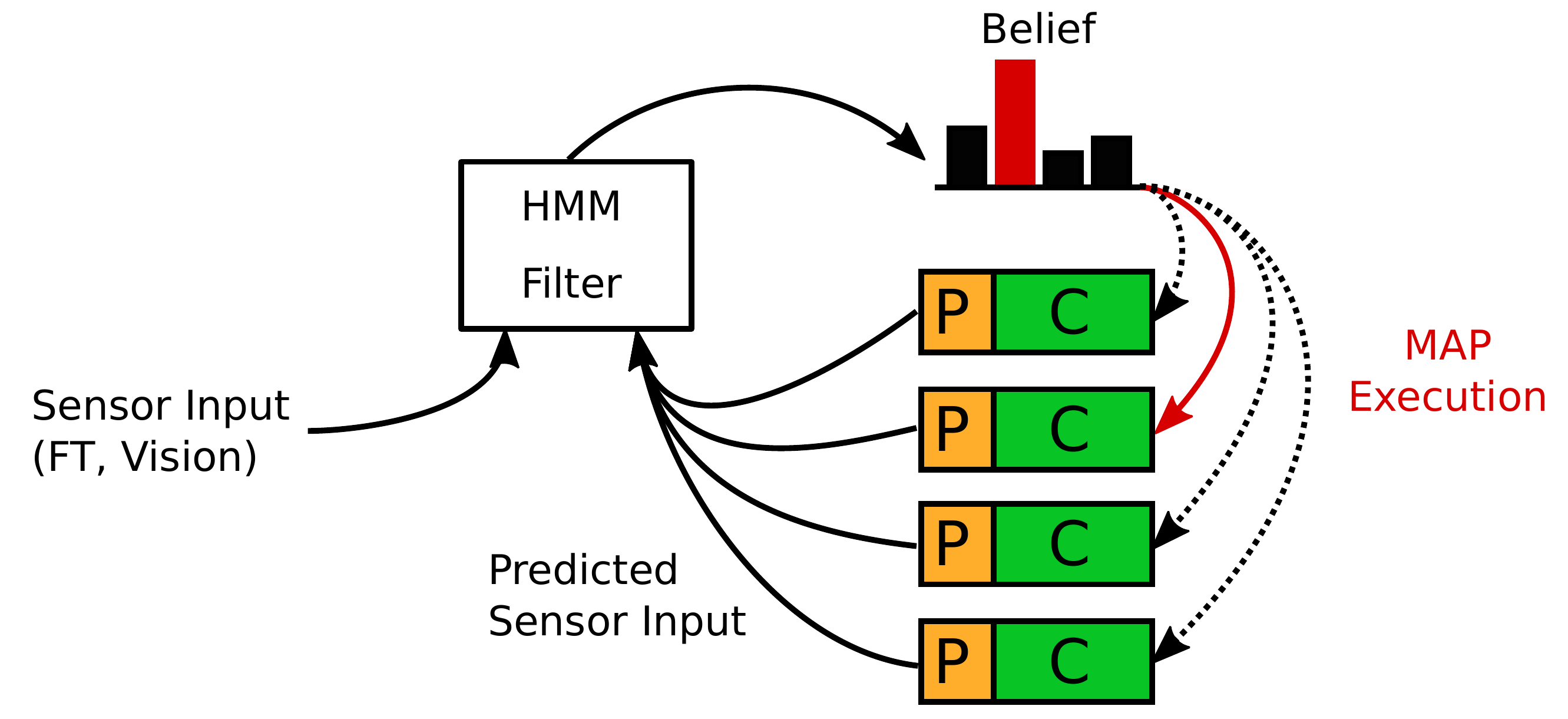}
 \caption{A schematic of our approach. FT and Vision input are integrated into a belief using Bayesian inference. The controller that corresponds to the discrete maximum a-posteriori (MAP) state is being executed.}\label{fig:method:approach}
\vspace{-0.25cm}
\end{figure}

A HMM consists of a discrete-time, discrete-state Markov chain, with hidden states $x_t \in {1,...,K}$ that evolve according to a state-transition-probability matrix $A_{ij}=p(x_t=i|x_t=j)$, plus an observation model $p(z_t|x_t)$ that determines the likelihood of observations $z_t$~\cite{murphy_machine_2012}. The forward algorithm can be used to filter the belief $bel(x_t)$ by alternating a prediction step (Equation~\ref{eq:hmm:predict}) and an update step (Equation~\ref{eq:hmm:update}) as
\begin{align} 
\overline{bel}(x_t) &= p(x_t|x_{t-1}) \: bel(x_{t-1}) = A^t bel(x_{t-1}) \label{eq:hmm:predict}\\ 
bel(x_t) &=  \eta \: p(z_t|x_t) \: \overline{bel}(x_t), \label{eq:hmm:update} 
\end{align}
where $\eta$ is a normalization factor. 

In our model a HMM replaces the controllers' preconditions and each controller is associated to a discrete state of the HMM. We always execute that controller which corresponds to the maximum a-posteriori (MAP) state estimate.

One of our main objectives with this paper is to show that when disturbances occur, behavior that relies on assumptions about the temporal sequence of controllers will be less robust than behavior which does not rely on them. In principle, a HMM's state-transition matrix $A$ can incorporate prior knowledge about the anticipated sequence of states. A strongly structured matrix $A$ may be important if the system is weakly observable due to noise or perceptual aliasing. However, the rich, multimodal sensor input in manipulation and the robot's ability to shape that input by interactive perception~\cite{bohg_interactive_2017} can often make the system observable enough so that we do not need a richly structured state-transition matrix $A$. The matrices $A$ we use in this work are weakly structured to contain $A_{ii}=0.95$ on the diagonal and $A_{ij} = \frac{0.05}{K-1}$ in all off-diagonal entries.  This state transition matrix does not model any assumptions about the temporal evolution of the discrete state of the task, except that it changes slowly. 
\section{Concrete Instantiations for Opening and Grasping From a Drawer}\label{sec:method:instantiation}
In this Section we will explain how we instantiate the previously described scheme for behavior that opens a drawer in Section~\ref{sec:method:smr:opening} and for behavior that graps tennis balls from a drawer in Section~\ref{sec:method:smr:grasp}. As these behaviors share common components we will first explain these in the next section.
\subsection{Shared Components of the Example Behaviors}

The robot we use is a WAM manipulator that has a force-torque (FT) sensor on its wrist and an RBO soft-hand~\cite{deimel16-IJRR} without thumb as its end-effector (EE), as in Figure~\ref{fig:robot}. We mounted a forward facing realsense D435 rgbd sensor below the hand. The FT and vision input both are incorporated into the HMM using separate measurement models as follows.

\subsubsection{Force-Torque Measurement Model}
When a robot performs multi-stage manipulation tasks, then different discrete states are often characterized by different statistical distributions of forces and torques acting on the EE.  We can measure these using the FT sensor. As the FT signal $z_{ft} \in \mathbb{R}^6$ is continuous it is common to define the likelihoods of each state's observation model $p_i(z_{ft,t}|x_t)$ as multivariate normal distributions. In Sections~\ref{sec:method:smr:opening} and~\ref{sec:method:smr:grasp} we will describe different expected FT-distributions as they arise during opening of a drawer and grasping tennis balls.  Crucially, the FT-signal strongly depends on both, the combined state of robot and environment, and the robot's actions. This makes it possible to perform interactive perception and to excert additional forces on the environment to reveal task-relevant information. This facilitates state estimation for the HMM and avoids perceptual aliasing.
\subsubsection{Vision Measurement Model}\label{sec:method:visionsvm}
Different states in a manipulation task do not only yield different FT distributions, but also different visual input $z_{vis}$.  Compared to FT measurements though it is difficult to define the likelihood as a generative model which predicts sensor input given state. It is more practical to learn a discriminative model for state given measurement~\cite{haarnoja_backprop_2016}.  In our case this means we learn a classifier to predict the likelihood $l_{vis,t}=Y(z_{vis,t})$ which we incorporate into the belief using Bayes rule as in $bel(x_t|z_{vis,t}) = \eta \: l_{vis,t} \: bel(x_{t-1})$. We use an SVM (RBF kernel, $C=1$) as a classifier on a bag-of-SIFT-features image descriptor~\cite{yang_comparing_2008}. 

This discriminative measurement model based on vision yields complementary information to the FT-based measurement model we described above. In Section~\ref{sec:method:smr:opening} we will now describe how these models are used to structure behavior when opening a drawer.

\subsection{Opening a Drawer}\label{sec:method:smr:opening}

\begin{figure}
\vspace{0.15cm}
 \includegraphics[width=\linewidth]{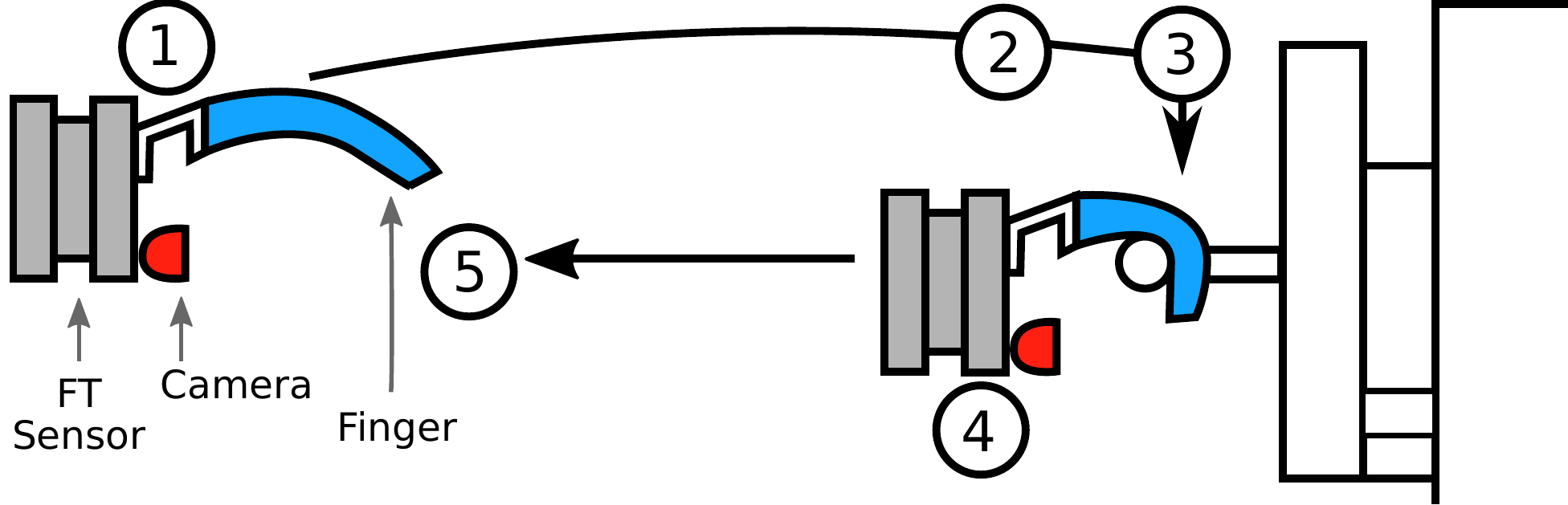}
 \caption{The discrete states in drawer opening as described in Section~\ref{sec:method:smr:opening}: (1) \emph{free-space}, (2) \emph{pre-grasp}, (3) \emph{front-plate contact}, (4) \emph{handle grasp}, and (5) \emph{drawer fully opened}}\label{fig:method:open}
\end{figure}

We will now now describe the discrete states we anticipate to occur while opening a drawer, which are also depicted in Figure~\ref{fig:method:open}. We explain the discrete high-level states in the order that corresponds to an idealized linear task structure, but in reality this sequence may not occur. It is not explicitly built into our system, as we previously explained in Section~\ref{sec:method:stateestimation}. For each discrete state we describe the expected sensor input the robot would receive and the controller that is executed when the robot is assumed to be in that state.

\subsubsection{Discrete State: Free Space}

In this phase the EE is in free space and the robot can see the drawer using its wrist-mounted rgbd camera. The FT input is expected to be distributed as $\vec{F}\sim\mathcal{N}(\vec{0}, I_6(\frac{1}{5},\frac{1}{5},\frac{1}{5},\frac{1}{20},\frac{1}{20},\frac{1}{20}))$, as we assume there is no contact with the environment. In the following we will just report parameters where the expected FT input differs from this distribution.

The likelihood of the visual measurement is estimated using an SVM, as described in Section~\ref{sec:method:visionsvm}. The training examples for this class are pre-recorded rgbd images of different views onto the drawer from a distance of $\approx 1m$.

In this situation we execute an image based visual servoing~\cite{chaumette_visual_2006} controller that moves the camera (and respectively also the end-effector) towards a pre-recorded view. In the desired viewing pose the end-effector is directly positioned in front of the drawer's handle. This relative pose between drawer and end-effector corresponds to a suitable pre-grasp pose. The servoing uses SIFT features matched with a ratio test threshold of $0.6$ and we use a table to memorize the last observed 3d position of features so that this controller can even visually servo back to a view on the drawer when it lost direct sight of it.

\subsubsection{Discrete State: Pre-grasp}

In this state the end-effector is positioned in a pre-grasp pose where it just needs to move forward to either directly grasp the handle or touch the drawer's front plate above the handle.

This phase is again characterized by the same expected FT distribution as before, but the training camera images are recorded in a small sphere around the pre-recorded view of the previous discrete state \emph{free-Space}. 

When this regularity is active we execute an impedance controller to move the end-effector forward in a straight line towards the drawer.

\subsubsection{Discrete State: Front-Plate Contact}

In this state we assume that the robot actively pushes against the front-plate of the drawer with its end-effector. We model this as an expected $F_z=-5N$. The training rgb-d views are recorded by moving the end-effector in a disk-like shape, above the handle and touching the front-plate.

In this state the robot activates an impedance controller that slides downwards in the $x$-direction of the end-effector frame, which is assumed to be a movement towards the handle of the drawer.  Importantly, the robot also servos to the expected sensed force of $F_z=-5N$, which means it pushes against the drawer's front plate.  This not only ensures that the end-effector slides onto the handle for a robust grasp, but it also maintains the sensed force $F_z$ that is expected as input when this system is in this state.  This is an example of interactive perception~\cite{bohg_interactive_2017} where the robot's behavior is adapted to facilitate the state estimation problem. Without this additional interactive force it would be difficult to distinguish this state from the \emph{pre-grasp} state, as both yield almost the same visual input and potentially also similar FT input (perceptual aliasing). 

\subsubsection{Discrete State: Handle Grasped}

In this state the robot pushes down onto the drawer's handle, potentially after it slid down the front-plate of the drawer.  The expected FT input is $F_x=-5N, \:T_y=-0.1Nm$, where $T_y$ is the torque around the end-effector's $y$-axis. The expected visual input is the same as in the previous discrete state \emph{Front-Plate Contact}

In this state we inflate the soft-hand to grasp the handle and execute an impedance controller that moves the end-effector in $-z$ direction to pull open the drawer. As in the previous state we again perform Interactive Perception to maintain the force of $F_z=-5N$ between end-effector and handle.  This idea goes back to a previous study~\cite{baum_achieving_2017} where we showed that we can facilitate the estimation of handle grasping success by actively maintaining such contact forces.

When the robot indeed has a robust grasp on the handle of the drawer the executed controller will converge towards a state where the drawer is opened.

\subsubsection{Discrete State: Drawer Fully Opened}

In this state the robot has grasped the handle and fully opened the drawer by pulling the handle.  As the robot has established a successful grasp and pushes down onto the handle while pulling on it at the drawer's joint limit, the expected FT input is $F_x=-5N, F_z=5N, T_y=-0.1Nm$. The visual input is the same as in the previous discrete state \emph{front-plate contact}.

In this state we deflate the soft-hand to ungrasp the handle and continue to pull back the end-effector, as in the previous state \emph{handle grasped} but without servoing towards a desired force in $x$-direction.

In Section~\ref{sec:eval:opening} we will show that the robot robustly opens a drawer while it transitions through the discrete states described in this section.  We will show that the interactively shaped feedback from the environment is sufficient to structure the robot's behavior. It chooses the appropriate controllers for execution, even when severe disturbances are present. In the next section we are going to explain the high-level states, controllers and estimators that are relevant when this robot grasps tennis balls from a drawer.

\subsection{Grasping From a Drawer}\label{sec:method:smr:grasp}

\begin{figure}
\vspace{.1cm}
 \includegraphics[width=\linewidth]{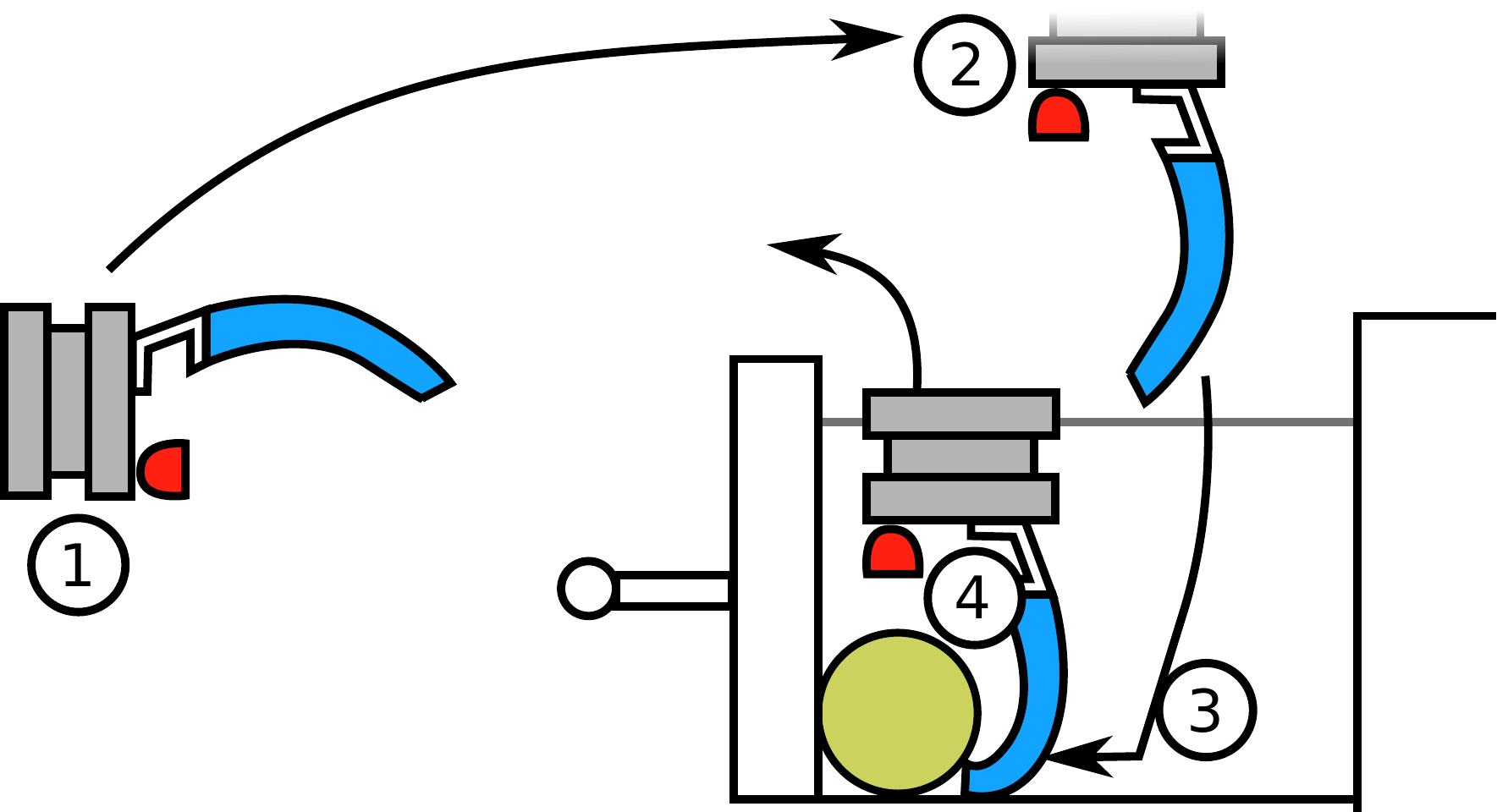}
 \caption{The discrete states in grasping tennis balls from a drawer as described in Section~\ref{sec:method:smr:opening}: (1) \emph{free-space}, (2) \emph{top-down view}, (3) \emph{pushing down}, (4) \emph{pushing against front plate}}\label{fig:method:grasp}
\end{figure}

In this task the robot's end-effector and camera are positioned in front of an opened drawer.  Inside that drawer there are three tennis-balls that the robot should remove.  The robot can solve this task by moving its camera above the drawer, detecting the balls and lifting them out of the container.  We will now explain the discrete states of that regularity in detail. They will again be ordered in an idealized sequence, although this sequence is not explicitly built into the system and will emerge by interaction with the environment.

\subsubsection{Discrete States: Views Onto the Drawer in Free-Space}

In this sub-section we explain three states that are very similar.  In each of these three states the robot can perceive a different view onto the opened drawer.  Once directly from the front, once from a little higher, looking diagonal into the drawer and another view that is even higher where the robot's camera diagonally points into the drawer at a steep angle.  The likelihood of these views is again computed by an SVM as described above and training data was pre-recorded in small spheres around those views.

Force-torque measurements are assumed to be distributed according to $\vec{F}\sim\mathcal{N}(\vec{0}, I_6(2,2,2,\frac{1}{2},\frac{1}{2},\frac{1}{2}))$, with significantly higher covariance than in the previous task. We chose a higher covariance because the FT signal in the contact states of this grasping task is less predictable.

The controllers executed in these states are visual servoing controllers that servo towards the pre-recorded next higher view into the drawer.  The top-most view activates a visual servoing controller towards a vertical top-down view into the drawer. Although there is no explicit sequence enforced, this combination of view classification and visual servo controller activation represents a sequence of \emph{funnels}~\cite{burridge_sequential_1999} that guide the robot's camera towards a top-down view into the drawer.

\subsubsection{Discrete State: Top-Down View Into the Drawer}
This state is characterized by the same expected FT input as before and the training RGB-D views are pre-recorded top-down views into the drawer from different distances and with tennis balls at different locations.

In this state the robot executes an impedance controller that moves the end-effector downwards into the drawer. The orientation of the end-effector is constrained, so that it always points downward. The $x$ and $y$ positions are controlled using a 2D visual servo controller, so that the end-effector is always dragged towards a position above the closest tennis ball. Tennis balls are detected using a classic computer vision approach, where the input image is first backprojected using pre-recorded image patches of tennis balls, thresholded, then erode-dilated and finally blob-detection is applied.  As long as a tennis ball is in view of the camera, this controller can converge towards a pose above that ball where ultimately the end-effector pushes down onto the ball or the drawer's bottom plate.

\subsubsection{Discrete State: Pushing Down onto the Drawer's Bottom Plate or a Tennis Ball}

This state is characterized by the same expected visual input as the state before, but with an expected FT input $F_z=-5N$. This force occurs as the EE pushes downward against the drawer's bottom plate or a ball.

In this state we inflate the soft-hand halfway and execute an impedance controller that moves in the end-effector's $x$-direction, towards the front-plate of the drawer.  Additionally, the impedance controller servos towards the expected input force $F_z=-5N$, which helps the state estimation assign a high likelihood to this discrete state. This a similar example for interactive perception as in the previous task.

While this controller is active, the robot continuously moves towards the front-plate of the drawer until it either pushes directly against the that plate or until it presses a tennis ball against it.

\subsubsection{Discrete State: Pushing Against the Drawer's Front Plate or a Tennis Ball}

This state is characterized by the same expected visual input as before, but with a different expected FT input $\vec{F}\sim\mathcal{N}((-5N, 0, 0, 0, 0, 0), I_6(2,2,150,\frac{1}{2},\frac{1}{2},\frac{1}{2}))$. Here we expect a force that pushes against the end-effector from the end-effector's $x$-axis, because the end-effector pushes against the drawer's front-plate. We assign a high-variance to the expected $F_z$, as the force $f_z$ in this state is hardly predictable.

In this state we fully inflate the hand, so that it graps a tennis ball if there is one in front of the hand. We further activate an impedance controller that lifts the hand out of the drawer. To facilitate state recognition, this controller also servos to a desired force of $F_x=-5N$, pushing against the front-plate. This not only serves for interactive perception, but also facilitates the grasp, as it further pushes a tennis ball that may be in front of the hand into a stable grasp.

In Section~\ref{sec:eval:grasp} we will show that the robot is able to robustly grasp tennis balls from the drawer, while it transitions through the discrete states described in this section.  We will show that feedback from the environment is sufficient to properly structure the robot's behavior, so that it chooses the appropriate controllers to be executed, even when severe disturbances are present.
\section{EXPERIMENTAL EVALUATION}\label{sec:experiments}

Our main concern is to show that the generated behavior can be robust to severe and unforeseen disturbances that may challenge temporal assumptions about the tasks. To this end, we evaluate how the robot's success rate evolves under increasing levels of interference.  We make this analysis in two different settings: In Section~\ref{sec:eval:opening} we evaluate performance in opening a drawer, and in Section~\ref{sec:eval:grasp} we evaluate performance in grasping tennis balls from a drawer.  We compare our method to a baseline that is a sequential Hybrid Automaton~\cite{egerstedt_behavior_2000} which uses the same controllers as ours, but which assumes a linear sequence of those controllers in the order they were described in the previous sections.  The baseline switches to the next node when the likelihood of that node according to the Hadamard product of FT and Vision likelihood is higher than that of the current mode.

\subsection{Removing Temporal Assumptions Increases Robustness When Opening a Drawer}\label{sec:eval:opening}

\begin{figure}
 \includegraphics[width=\linewidth]{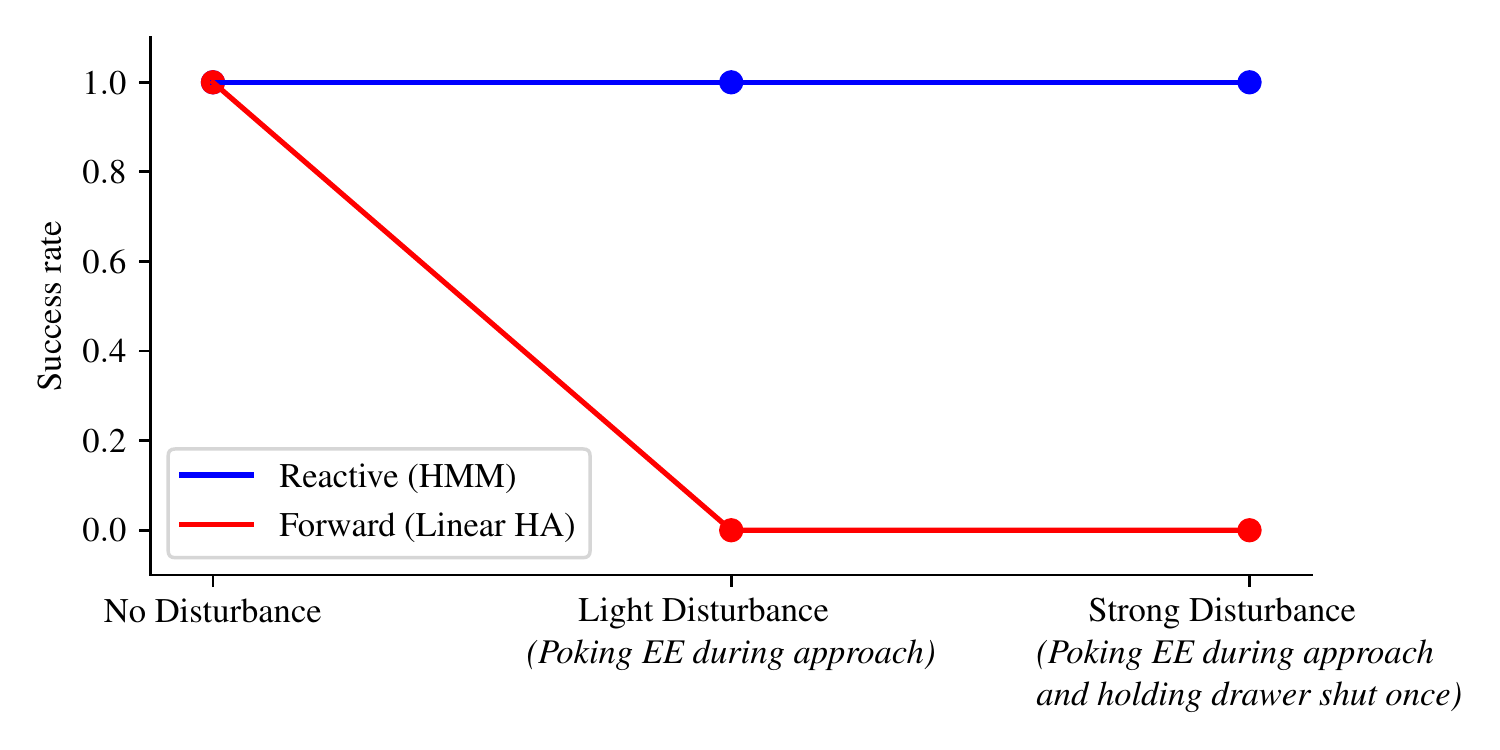}
 \caption{Success rate in drawer opening plotted for experimental conditions with increasingly severe interferences. Each condition is repeated $n=10$ times. The linear baseline and our reactive approach succeed reliably in abscence of disturbances, however only the reactive method is robust to disturbances as it relies less on temporal assumptions.}\label{fig:eval:open}
\end{figure}

To understand how robust the behavior is to unforeseen interferences, we tested it under several conditions that are increasingly challenging. Figure~\ref{fig:eval:open} shows success rate plotted over these conditions. To compute the success rate we ran the experiment $n=10$ times from different starting conditions and recorded how often the robot was able to open the drawer. If there was no disturbance our method and the baseline were both able to open the drawer 10/10 times. 

In the \emph{Light Disturbance} condition we pushed against the EE during the approach phase, which triggered the robot to shift its belief towards states that assume contact with the drawer, which \emph{ideally} only happens after the approach phase. Our approach (\emph{Reactive}) still opened the drawer 10/10 times while the baseline (\emph{Linear}) was never able to recover because its assumed temporal structure was violated.

In the \emph{Strong Disturbance} condition a person first performed the same interference as in the \emph{Light Disturbance} condition, and then also kept the drawer shut the first time that the robot established a grasp and tried to pull open the drawer. Again our method could recover and solve the task 10/10 times while the baseline never succeeded. 

These results clearly show that the behavior becomes significantly more robust when we remove temporal assumptions and instead use feedback to structure behavior.

\begin{figure}
 \includegraphics[width=\linewidth]{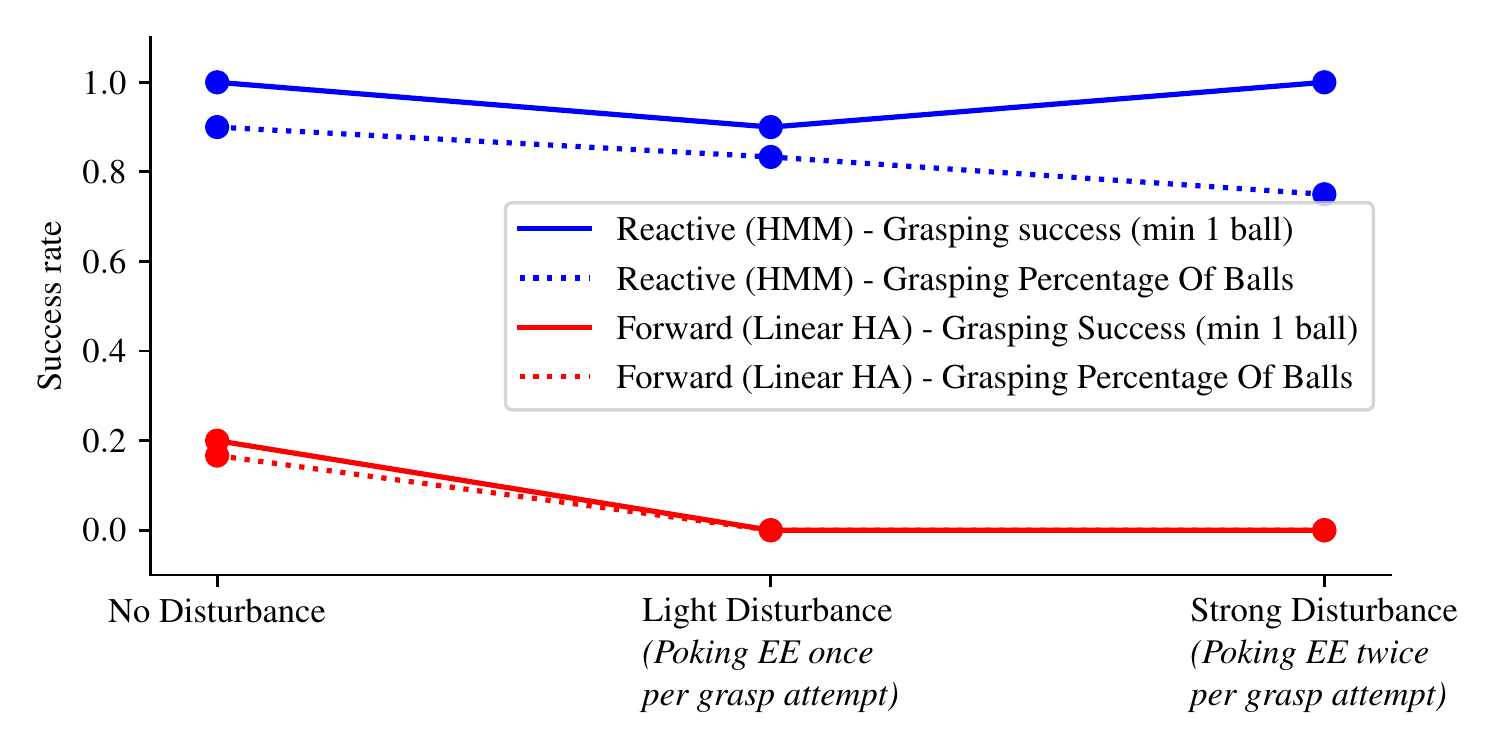}
 \caption{Success rate in grasping balls from a drawer plotted for experimental conditions with increasingly severe interferences. Each condition is repeated $n=10$ times. Because grasping and positioning errors occur regularly, the task structure is hardly predictable and not linear. This is challenging for the linear baseline, but the reactive method can reliably solve this task, even when interference occurs.}\label{fig:eval:grasp}
\end{figure}

\subsection{Removing Temporal Assumptions Increases Robustness When Grasping Tennis Balls}\label{sec:eval:grasp}

This experiment starts in a state where the EE and camera are positioned in front of an open drawer's front plate.  It has to grasp tennis balls, filled with sand, out of the drawer. We evaluate the robot's success rate in a condition with no outside disturbance and under conditions with increasing interventions from a human. In the \emph{Light Disturbance} conditions we pushed once against the EE for each one of the three balls to be grasped. In the \emph{Strong Disturbance} condition this interference happened two times.

In Figure~\ref{fig:eval:open} we plot two measures of task performance over the different experimental conditions.  We evaluate how often the robot could grasp at least one ball from the drawer per experiment and plot that as solid lines.  The dashed lines represent how many of the maximum amount of balls the robot could remove, normalized by this maximum number. A value of $0$ would mean the robot could not remove any balls on average and a value of $1$ means the robot was able to remove all 3 balls on average.

The data shows that the HMM-based approach maintains high performance irrespective of interferences while the baseline largely could not succeed at all. The baseline also had low performance in the \emph{No Disturbance} condition, because even without interference the task-structure in the grasping task is more unpredictable than a linear plan could cope with. This is because often balls slip out of the hand and need to be regrasped, but then the \emph{Top} view would have to be re-established to localize a ball and start a new loop. 

These results again show that the behavior becomes significantly more robust when we remove temporal assumptions.

\section{CONCLUSION}

We showed that even simple manipulation systems, using standard tools from state estimation and control, can create suprisingly robust behavior.
We designed two reactive behaviors to open a drawer and grasp tennis balls from it and evaluated their task performance under increasing levels of disturbance. Both were extremely robust, even against significant outside interventions.
Robustness was enabled because the behaviors did not rely on assumptions about the sequence of controllers to activate, but they chose controllers online and performed interactive perception to boost the information content in sensor input.
We conclude that behavior can be more robust when it relies on feedback instead of assumptions about temporal task structure, as far as this is possible. The quality of that feedback is crucial and interactive perception is the right paradigm to enable informative feedback.

\balance
\bibliography{references}

\end{document}